# Deep learning based cloud detection for medium and high resolution remote sensing images of different sensors

Zhiwei Li [a], Huanfeng Shen [a,b,c,*], Qing Cheng [d,*], Yuhao Liu [a],

Shucheng You [e], Zongyi He [a]

[a] School of Resource and Environmental Sciences, Wuhan University, Wuhan, China.

[b] Collaborative Innovation Center of Geospatial Technology, Wuhan, China.

[c] Key Laboratory of Geographic Information System, Ministry of Education, Wuhan University, Wuhan, China.

[d] School of Urban Design, Wuhan University, Wuhan, China.

[e] Department of Remote Sensing, China Land Surveying and Planning Institute, Beijing, China.

[*] Corresponding authors.

E-mail address: shenhf@whu.edu.cn (H. Shen), qingcheng@whu.edu.cn (Q. Cheng).

## ABSTRACT

Cloud detection is an important preprocessing step for the precise application of optical satellite imagery. In this paper, we propose a deep learning based cloud detection method named multi-scale convolutional feature fusion (MSCFF) for remote sensing images of different sensors. In the network architecture of MSCFF, the symmetric encoder-decoder module, which provides both local and global context by densifying feature maps with trainable convolutional filter banks, is utilized to extract multi-scale and high-level spatial features. The feature maps of multiple scales are then up-sampled and concatenated, and a novel multi-scale feature fusion module is designed to fuse the features of different scales for the output. The two output feature maps of the network are cloud and cloud shadow maps, which are in turn fed to binary classifiers outside the model to obtain the final cloud and cloud shadow mask. The MSCFF method was validated on hundreds of globally distributed optical satellite images, with spatial resolutions ranging from 0.5 to 50 m, including Landsat-5/7/8, Gaofen-1/2/4, Sentinel-2, Ziyuan-3, CBERS-04, Huanjing-1, and collected high-resolution images exported from Google Earth. The experimental results show that MSCFF achieves a higher accuracy than the traditional rule-based cloud detection methods and the state-of-the-art deep learning models, especially in bright surface covered areas. The effectiveness of MSCFF means that it has great promise for the practical application of cloud detection for multiple types of medium and high-resolution remote sensing images. Our established global high-resolution cloud detection validation dataset has been made available online (http://sendimage.whu.edu.cn/en/mscff/).

**Keywords**: Cloud detection; Cloud shadow; Convolutional neural network; Multi-scale; Convolutional feature fusion; MSCFF

**1. Introduction**

Clouds in optical remote sensing images are inevitable and limit the potential of the imagery for ground information extraction, and thus the existence of clouds and the accompanying cloud shadows affects the availability of satellite data. Cloud and cloud shadow detection is essential before the precise application of satellite images. Accurate cloud extraction from cloudy images is essential to reduce the negative impact of clouds on image applications (Le Hégarat-Mascle and André, 2009; Zhu and Woodcock, 2014; Lin et al., 2015; Wu et al., 2016). Therefore, cloud detection in optical satellite images is of great significance.

In recent years, scholars have done a great deal of research into cloud detection for different types of satellite images, including Landsat (Irish et al., 2006; Zhu and Woodcock, 2012; Braaten et al., 2015; Zhu and Helmer, 2018), the Moderate Resolution Imaging Spectroradiometer (MODIS) (Platnick et al., 2003; Liu et al., 2004; Luo et al., 2008; Frey et al., 2008; Ishida et al., 2018), Sentinel (Hagolle et al., 2010; Frantz et al., 2018), and the Medium Resolution Imaging Spectrometer (MERIS) (Gómez-Chova et al., 2007; Mei et al., 2017). Most of the current cloud detection methods extract the clouds from the imagery through rule-based classification, which is based on the physical properties of cloud. This category of rule-based methods is often used to deal with a particular type of satellite imagery. The most well-known cloud detection method for Landsat imagery is Fmask (Zhu and Woodcock, 2012; Zhu et al., 2015), which is a rule-based method utilizing multiple threshold tests based on cloud physical characteristics, which can achieve a high accuracy of cloud detection in most circumstances. Considering the defects of the Fmask method in mountainous areas and bright land surfaces,

Qiu et al. (2017) and Frantz et al. (2018) improved Fmask with the aid of digital elevation models (DEMs) and by presenting a cloud displacement index, respectively. In addition, Luo et al. (2008) employed a scene-dependent threshold-based decision matrix to identify clouds in MODIS imagery for clear-sky image composition. Sedano et al. (2011) coupled pixel-based cloud seed identification and object-based cloud region growing to extract the clouds in SPOT imagery. To reduce the effects of atmospheric factors and complex surfaces, an *a priori* monthly surface reflectance database was established in Sun et al. (2016) to support the universal dynamic threshold cloud detection algorithm (UDTCDA) algorithm for cloud detection in MODIS and Landsat-8 imagery. Zhai et al. (2018) later proposed a unified method for cloud and cloud shadow detection in multi/hyperspectral images, based on spectral indices and spatial matching, with parameters which may need to be fine-tuned. Multi-temporal cloud detection methods (Hagolle et al., 2010; Goodwin et al., 2013; Zhu and Woodcock, 2014; Lin et al., 2015; Bian et al., 2016; Mateo-García et al., 2018; Zhu and Helmer, 2018) involving temporal information have also been employed for time-series imagery, to further improve cloud detection accuracy.

**Table 1** Selected cloud detection methods for different optical satellite imagery.

| Authors | Date | Method name | Applicable imagery (mainly) |
| --- | --- | --- | --- |
| Simpson et al. | 1998 | SMC | ASTR |
| Di Vittorio and Emery | 2002 | DTCM | AVHRR |
| Gómez-Chova et al. | 2007 | N/A | ENVISAT MERIS |
| Luo et al. | 2008 | CCRS | MODIS |
| Zhu and Woodcock | 2012 | Fmask | Landsat TM, ETM+, OLI/TIRS |
| Fisher | 2014 | SPOTCASM | SPOT-5 HRG |
| Braaten et al. | 2015 | MSScvm | Landsat MSS |
| Harb et al. | 2016 | CSDT | CBERS HRCC, Landsat TM/ETM+ |
| Bian et al. | 2016 | N/A | HJ-1A/B CCD |
| Li et al. | 2017 | MFC | Gaofen-1 WFV |
| Qiu et al. | 2017 | MFmask | Landsat-4-8 |
| Mei et al. | 2017 | XBAER-CM | ENVISAT MERIS |
| Frantz et al. | 2018 | N/A | Sentinel-2 |
| Sun et al. | 2018 | N/A | Landsat-8 |

However, there are still some problems in cloud detection for satellite imagery which have not yet been solved well. Table 1 lists selected cloud detection methods developed in recent years for different types of optical imagery, from which it can be seen that most of the methods are designed for specific satellite imagery. With the increase of satellite image sources, it is inefficient to develop different methods for different satellite images. The application requirements of cloud detection for multi-source satellite images prompts us to propose a more general cloud detection method.

In addition, the traditional rule-based cloud detection methods often suffer from bright non-cloud object commission and thin cloud omission, especially for images with limited spectral information (Li et al., 2017, 2018). In order to further improve the cloud detection accuracy for a single image, more spatial features, such as geometric and texture features, can be combined with the spectral features to enhance the feature diversity (Bai et al., 2016; Li et al., 2017). However, since most of the cloud detection methods proposed in the previous studies only use low-level spectral and spatial features, there is still room for promoting cloud detection accuracy by using features at higher levels (Li et al., 2018). In addition, the input to the cloud detection should ideally be top of atmosphere (TOA) reflectance rather than digital number values, in most circumstances, to reduce the radiation difference between different images (Choi and Bindschadler, 2004; Irish et al., 2006; Gómez-Chova et al., 2007; Zhu and Woodcock, 2012; Li et al., 2017). Nevertheless, radiometric calibration is not always accurately implemented on satellite images, especially for sensors that lack onboard calibration capacity. The cloud detection for such imagery is thus more challenging.

To better deal with the above problems of the expansibility and accuracy of the current cloud detection methods, machine learning techniques for cloud detection have been introduced. The machine learning based methods mark the clouds in satellite imagery by a pre-trained model, which requires cloudy images and the corresponding cloud labels for model parameter learning. Hughes and Hayes (2014) proposed the spatial procedures for automated removal of cloud and shadow (SPARCS) algorithm, which trains a neural network to identify the cloud and cloud shadow in Landsat scenes. In order to cope with the highly varying patterns of clouds, Yuan and Hu (2015) constructed a bag-of-words model based on segmented super-pixels, and applied a support vector machine (SVM) classifier to discriminate cloud and non-cloud regions. To better generalize a cloud detection method with adaptability to various conditions, even with limited training samples, Bai et al. (2016) and Ishida et al. (2018) incorporated SVM classification to implement cloud detection for high-resolution satellite images and MODIS data, respectively.

As a subset of machine learning, deep learning has created huge breakthroughs, and is now taking off in the remote sensing field (Zhang et al., 2016; Zhu et al., 2017; Mountrakis et al., 2018). Benefiting from the application of deep convolutional features, the methods based on deep learning have achieved high accuracies in image classification tasks, and the accuracy is continuously being promoted with the development of new techniques (Li et al., 2018). Deep learning has also been introduced to cloud detection in satellite imagery in recent studies. In the study of Mateo-García et al. (2017), a simple convolutional neural network (CNN) architecture was designed for the cloud masking of Proba-V multispectral images. The experimental results suggested that CNNs have promise for solving cloud masking problems, compared to the classical machine learning approach. Another scheme was provided in Xie et al. (2017), in which the images are first segmented, and then the clouds in patches of the different segmented regions are labeled by the trained CNN network. Zhan et al. (2017) also applied a deep convolutional network to distinguish cloud and snow from satellite images at the pixel level.

Although the previous deep learning based cloud detection methods can acquire a relatively high accuracy, most of the methods have only been validated in local regions on a specific type of imagery. In this paper, we present a deep learning based multi-scale convolutional feature fusion (MSCFF) method to extract clouds and cloud shadows from multiple types of satellite imagery, which were collected from global regions in various land-cover types. The experimental results indicate that the proposed MSCFF method is effective and performs better than the compared methods in distinguishing clouds and bright land surfaces, as well as generating a more refined cloud and cloud shadow mask.

The rest of this paper is organized as follow. Section 2 introduces the proposed method and provides the implementation details. The experimental data are described in Section 3. The accuracy and efficiency of MSCFF and the compared methods are evaluated in Section 4, in which multiple types of imagery are used for the method validation. In Section 5, we describe the application extension of the proposed method to more types of imagery, and then discuss the limitations of the proposed method. Our conclusions are drawn in Section 6.

## 2. Methodology

The MSCFF method of cloud and cloud shadow detection is proposed to generate masks for multiple types of satellite imagery. As shown in Fig. 1, the flowchart of the proposed method is made up of two stages: model training and model testing. The experimental data described in the following section were divided into training data and test data for the model training and testing, respectively.

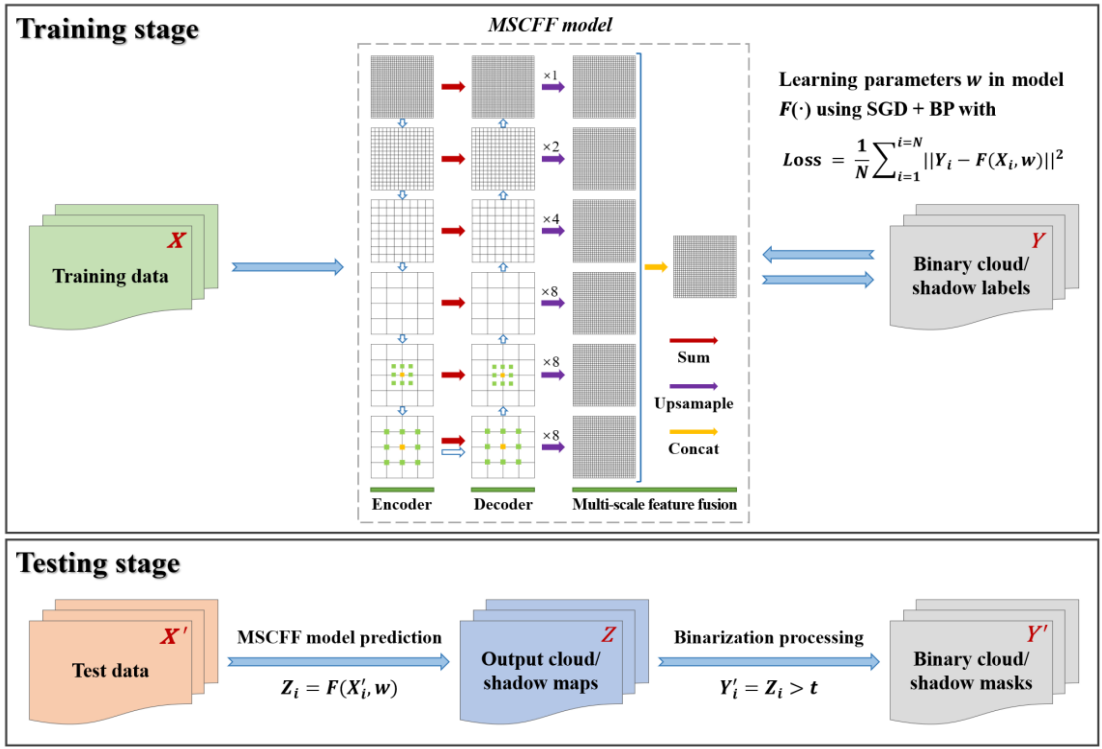

**Fig. 1.** Flowchart of cloud and cloud shadow detection based on the multi-scale convolutional feature fusion method.

In the model training stage, the images and the corresponding cloud and cloud shadow labels for MSCFF model training are first preprocessed, and then the parameters in the MSCFF model can be iteratively learned to reach an optimal allocation by minimizing the loss function, which is computed over the model predictions and binary labels. The MSCFF model is made up of a symmetric encoder-decoder module and a multi-scale feature fusion module, which are used for multi-scale feature extraction and fusion, respectively. The two output feature maps of MSCFF are cloud and cloud shadow maps.

In the model testing stage, in which the optimal parameters for the model are acquired, the test images used for the model testing are preprocessed, and then the pre-trained MSCFF model is used for cloud and cloud shadow prediction for each pixel in the image. The two output feature maps of cloud and cloud shadow for each image are fed to binary classifiers outside the model for pixel-wise binarization processing, and finally merged into a single cloud and cloud shadow mask. For the test images which contain cloud and cloud shadow labels, quantitative evaluation of the model performance can be conducted. The details of the data preprocessing and the proposed MSCFF method are provided in the following subsections.

**2.1. Data preprocessing**

In the model training stage, before using an image and its corresponding cloud and cloud shadow mask as samples for model training, we normalize the range of values of the image to [0,1] by maximum normalization or radiation calibration, depending on whether the image can be accurately calibrated. For imagery which can be accurately calibrated, such as Landsat, we normalize the image values to TOA reflectance by radiometric calibration. The maximum normalization method not only converts the imagery of different sources to the same value range, but also retains the initial spectral information in the imagery. The corresponding cloud mask and cloud shadow mask are separately generated, in which cloud or cloud shadow pixels are labeled as 1, while the others are set as 0. Note that the large multispectral image and mask are both clipped to a height and width of 256 × 256 to form a pair of training samples, where each sample does not overlap with any other. Since there are enough image patches for the model training, no data augmentation is utilized in the training stage.

In the model testing stage, the same image normalization method that was used in the training stage is again used. An entire image is divided into a series of equal patches for patch-by-patch processing, and each two neighboring patches are partially overlapped to avoid artifacts on the border of the classified patches. Specifically, the height and width of the input image patch can be 256 × 256, 512 × 512, 1024 × 1024, or a larger size, according to the memory capability of the device, while the number of input channels is dependent on the available bands in the experimental data.

**2.2. The multi-scale convolutional feature fusion method**

As shown in Fig. 2, the architecture of the proposed MSCFF method is a fully convolutional network, in which the output is a pixel-to-pixel map with the same height and width as the input. The fully convolutional network (FCN) (Shelhamer et al., 2017) is used for pixel-to-pixel image segmentation tasks, in which the fully connected layer, which connects every neuron in one layer to every neuron in another layer, and is typically used in CNNs for classification, is replaced by a convolutional layer to enable the network to classify each pixel in the image, instead of outputting a class label for the whole image. MSCFF differs from the previous deep cloud detection architectures (Mateo-García et al., 2017; Xie et al., 2017), which use fully connected layers to output a class label for the whole image patch. Specifically, as shown in Fig. 1, we utilize a symmetric encoder-decoder module in MSCFF to extract multi-scale and high-level features. Such a module has also been applied in SegNet (Badrinarayanan et al., 2017) for semantic pixel-wise segmentation. To make the network easy to train and more effective, the encoder-decoder module in MSCFF is improved by using residual network units (He et al., 2016) and dilated convolutions (Yu and Koltun, 2016; Chen et al., 2018). In addition, a multi-scale convolutional feature fusion module is designed and used in MSCFF to make full use of the convolutional features of different scales, and to further improve the accuracy of the cloud detection.

**2.2.1 Layers in the architecture of MSCFF**

The architecture of MSCFF is made up of two modules—an encoder-decoder module and a multi-scale feature fusion module—which both consist of a number of convolutional layers, pooling layers, and deconvolutional layers.

**A. The basic convolutional layer**

A convolutional layer is the most commonly used layer in MSCFF for feature extraction and feature fusion. The input of each convolutional layer is $m \times n \times c$ feature maps, where $m$ and $n$ are the height and width of the feature maps, respectively, and $c$ is the number of channels. Assuming that the convolutional layer has $k$ filters of size $r \times r \times c$, the output of the convolutional layer is feature maps with a size of $m' \times n' \times k$. We let $W_i$ and $b_i$ be the $i$-th filter and bias, and $X$ be the input feature maps. The output $i$-th feature map $Y_i$ can then be calculated by the equation $Y_i = W_i * X + b_i$, where $*$ is the discrete convolutional operator. In MSCFF, the output features maps of each convolutional layer, except the last, are batch normalized (Ioffe and Szegedy, 2015). Following this, an element-wise rectified linear unit (ReLU) of $f(x) = max(0, x)$ is applied. The sizes of the single convolutional filters are all 3 × 3. The height and width of the output feature maps of each convolutional layer are kept the same as the input, by padding zeros at the image borders before the convolution operation.

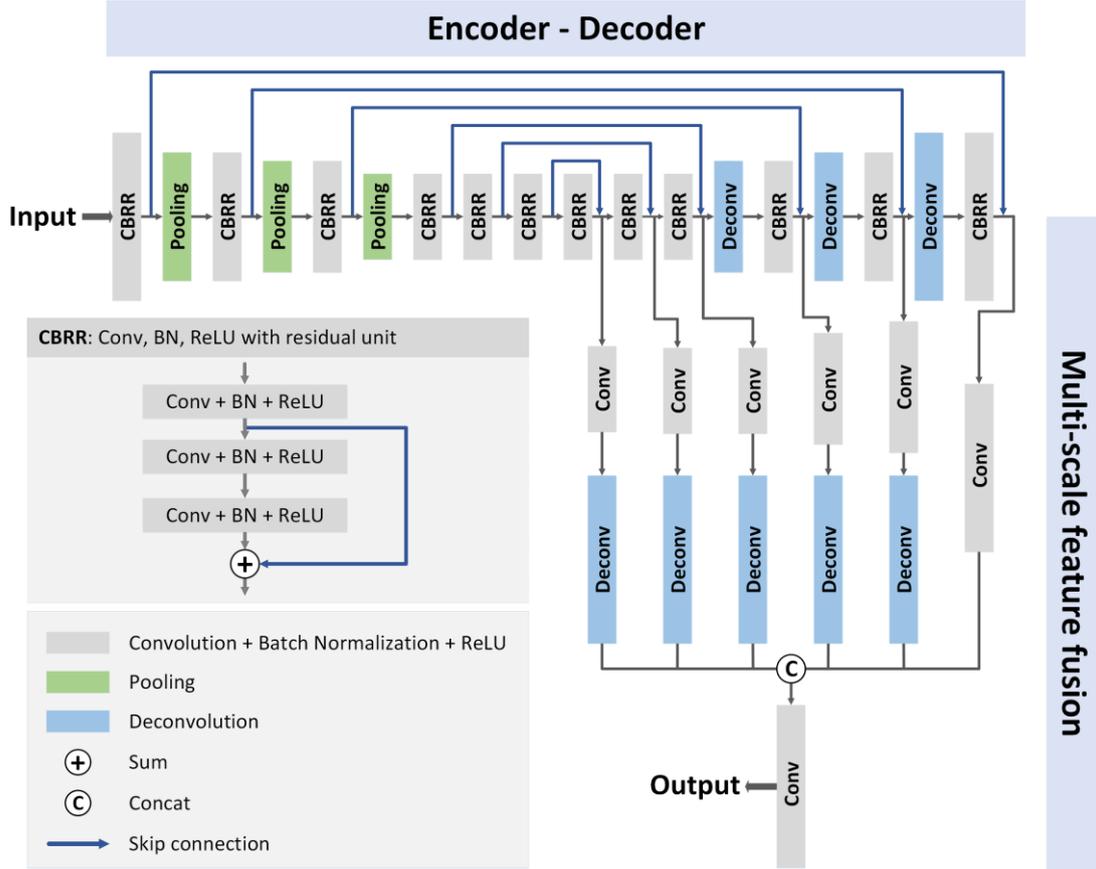

**Fig. 2.** The architecture of MSCFF. To simplify the illustration of the network structure, we refer to a block as a CBRR block, which contains three convolutional layers (Conv), batch normalizations (BN), and rectified linear units (ReLU) with a residual unit. The encoder-decoder module is utilized to extract multi-scale features, and then the up-sampled feature maps of six scales are concatenated and fused for the output. The two output feature maps of the last convolutional layer are the cloud and cloud shadow maps, in which the values are supposed to range from 0 to 1, which are then fed to binary classifiers outside the model for pixel-wise cloud and cloud shadow classification.

**B. The dilated convolutional layer**

Contextual information can effectively promote the accuracy of classification. Deep CNNs enhance the contextual information through enlarging the receptive field during the convolution operations. Here, the receptive field is defined as the region of the input space that affects a particular unit of the network, and the receptive field of a feature can be described by its center location and its size. The target of enlarging the receptive field can be reached by increasing the layers of the network and enlarging the size of the convolutional filter. However, as the network depth and filter size increase, this results in a significant increase in computational burden and training time. To solve this issue effectively, dilated convolutional layers are introduced in MSCFF, which support the exponential expansion of the receptive field while maintaining the filter size, by dilating the convolutional kernels. Yu and Koltun (2016) and Yu et al. (2017) first proposed dilated convolutions to systematically aggregate multi-scale contextual information, without losing resolution, and their experiments suggested that dilated networks outperform the non-dilated counterparts in image classification, without increasing the model depth or complexity. Differing from basic convolution, the dilated convolution operator can employ the same filter at different ranges using different dilation factors. Setting the kernel size as 3 × 3 and the kernel dilation as two pixels as an example, the receptive field of basic convolution has a linear correlation with the layer depth $d$, in that the receptive field size $R = (2d + 1) \times (2d + 1)$, while the dilated convolution receptive field has an exponential correlation with the layer depth, where the receptive field size $R = (2^{d+1} - 1) \times (2^{d+1} - 1)$ (Zhang et al., 2018). In MSCFF, the dilation factors of the 3 × 3 dilated convolutions in the last two CBRR blocks of the encoder and the first two corresponding CBRR blocks of the decoder are respectively set to 2, 4, 4, and 2. The increase of the receptive field size of the convolutional layer allows MSCFF to extract features from a larger field without changing the size of the feature map, thus preserving more spatial information for the multi-scale feature fusion and final pixel prediction.

**C. The pooling layer**

The pooling layer takes local regions of each feature map and extracts values for the spatial regions by maximum or average operators, to reduce the spatial dimensions, but not the depth. In MSCFF, we use three pooling layers with 2 × 2 filters and a two-pixel stride to down-sample the feature maps in the encoder by the maximum operator, which simply selects an input element for each output element. The reason why we only use pooling layers after the first three CBRR blocks, and use dilated convolutional layers instead of basic convolutional layers in the last two CBRR blocks in the encoder, is that the combined use of pooling layers and dilated convolutional layers in the encoder provides more strong features of multiple scales, as well as preserving more detail information for the following decoder, thus finally achieving a better performance.

**D. The deconvolutional layer**

The deconvolutional layers are utilized to up-sample and reconstruct the feature maps. Deconvolution is also termed transposed convolution or fractionally strided convolution, and it proceeds in the opposite direction to normal convolution (Noh et al., 2015; Dumoulin and Visin, 2018). In MSCFF, we use deconvolution as the decoding layer of a convolutional encoder, to recover the shape of the initial feature map in the encoder stage.

**2.2.2 Implementation details of MSCFF**

MSCFF is made up of an encoder-decoder module and a multi-scale feature fusion module. The encoder-decoder module is a symmetric architecture of convolutional encoders and corresponding decoders. The idea is to use features extracted at different scales to provide both local and global context, where the feature maps of the early encoding layers retain more spatial details, leading to more refined boundaries (Badrinarayanan et al., 2017). The encoder extracts multi-scale low-level spatial features of the input image by the convolutional layers, where the pooling layers are used to down-sample the feature maps. The decoder up-samples and reconstructs the feature maps by the deconvolutional layers, where the convolutional layers are used to extract the high-level semantic features. The feature maps of equal sizes in the encoding layers and corresponding decoding layers are connected by skip connections and element-wise summed. Note that skip connections are extra connections between nodes in different layers of a neural network that skip one or more layers. Such connections can help to retain more spatial details in the reconstructed feature maps produced in the decoder, and also make the training of very deep networks possible.

Feature maps closer to the output contain higher-level semantic features, which are critical and contribute more to the output of the network (Shelhamer et al., 2017). Therefore, the multi-scale feature fusion module aggregates the multi-scale features from the decoder stage by up-sampling the six-scale feature maps to the same spatial size as the input. The up-sampled feature maps are then concatenated and fed to the last convolutional layer, which fuses the multi-scale features and outputs two maps corresponding to cloud and cloud shadow. All the output feature maps of the convolutional layers, except for the last, are batch normalized (Ioffe and Szegedy, 2015) and activated by ReLU. Tricks for optimizing deep networks, including residual learning (He et al., 2016) and gradient clipping (Kim et al., 2016), are also utilized to boost the network convergence during training and improve the network performance.

To make the network focus on the generation of the cloud and cloud shadow maps for the following binarization processing, we use mean-squared error loss in MSCFF instead of cross-entropy loss, which is commonly used in similar classification tasks, to make the network only output cloud and cloud shadow maps, in which the values are supposed to range from 0 to 1. Note that the accuracies of MSCFF models using cross-entropy loss and mean-squared error loss are nearly equal, according to our evaluation, while the output of a model using mean-squared error loss is sometimes more convenient in actual applications, which motivated us to choose mean-squared error loss as the loss function. For example, a certain degree of cloud/cloud shadow commission is tolerable in land change surveys, to reduce omission errors as far as possible, and it is more convenient if we can intuitively balance the commission and omission errors for specific types of imagery to generate more desirable masks by adjusting the threshold to binarize the output cloud and cloud shadow maps of the MSCFF model.

Given a training dataset containing multispectral images $x_i$ and corresponding masks $y_i$, the goal is to learn a model $F$ that predicts the cloud and cloud shadow maps of an input image patch. The optimal parameter $w$ in $F$ can be learned by minimizing the mean-squared error loss $L(w)$, which is averaged over the training data and defined as shown in Eq. (1), where $N$ is the number of training samples. Stochastic gradient descent (SGD) (Bottou, 2010) and backpropagation (BP) are utilized to acquire the optimal parameter $w$ by minimizing the loss function. The mini-batch size is set to 10. The learning rate decay policy is chosen as *poly*, with an initial learning rate of 0.1, in which the learning rate follows a polynomial decay from 0.1 to zero when reaching the maximum iteration limit. Finally, the MSCFF model is trained from scratch for 200,000 iterations on MatConvNet (Vedaldi and Lenc, 2015), with support from an NVIDIA Titan Xp GPU. The model training took 2–3 days for each dataset listed in the experimental data section.

$$L(w) = \frac{1}{N}\sum_{i=1}^{i=N} ||y_i - F(x_i, w)||^2 \qquad (1)$$

In the test stage, as shown in Fig. 1, the output of the MSCFF model is cloud and cloud shadow maps. Due to the limitation of the device capability, the prediction for a whole image is produced patch by patch. In order to avoid artifacts on the border of the classified patches, each two neighboring patches are partially overlapped, and the maximal values are selected as the final predictions for the overlapped regions. For an output of cloud and cloud shadow maps, an appropriate threshold can be selected to binarize the maps to cloud and cloud shadow masks, according to the specific requirements. In this paper, a default segmentation parameter $t$ of 0.5 is chosen to better balance the commission and omission errors in the cloud and cloud shadow masks. Finally, the binary masks are merged into a single cloud and cloud shadow mask, in which a higher priority is set for cloud. Therefore, a pixel is labeled as cloud when it is also labeled as cloud shadow.

**2.2.3 Network architecture validation**

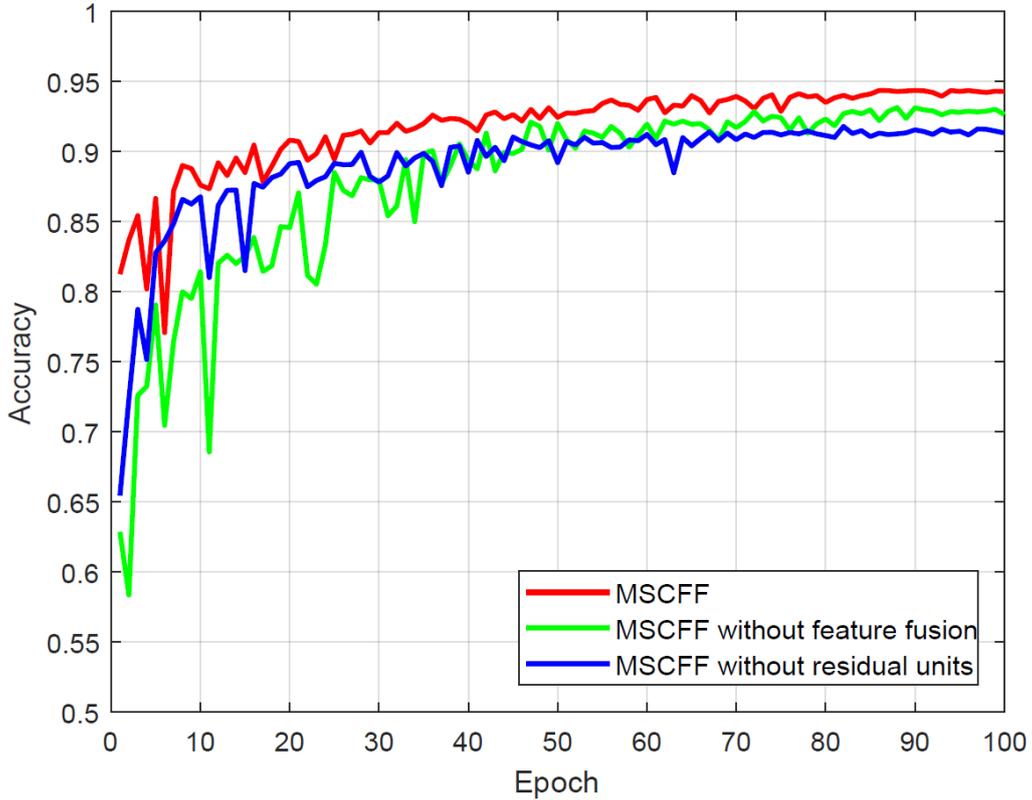

**Fig. 3.** Accuracy curves over a validation subset of our established high-resolution dataset for MSCFF, MSCFF without feature fusion, and MSCFF without residual units.

In order to validate the effectiveness of the network, we trained and tested the performance of MSCFF under different network configurations on training and validation subsets of our established high-resolution dataset, which is introduced in the following section. We compared the performances of MSCFF and models without feature fusion and without residual units, to demonstrate the benefits of the addition of the multi-scale feature fusion module and the residual network units in the encoder-decoder module. MSCFF without feature fusion denotes the network without the multi-scale convolutional feature fusion module, where we generate the output from the last convolutional layer in the decoder module. In the network of MSCFF without residual units, we remove the residual unit from each CBRR block shown in Fig. 2.

Except for the changes in the model structure, all the other parameter settings of MSCFF and the compared models were the same in the training and testing stages. The experimental results are shown in Fig. 3, from which we can see that both the multi-scale feature fusion module and the residual units in MSCFF can help to improve the cloud detection accuracy.

In addition to the model structure, the number of filters is a key factor affecting the model performance, and thus we compared the model performance of MSCFF with different filter settings. By default, we set the number of filters for both the convolutional and deconvolutional layers to 64 and, for comparison, we set the filter numbers to {64, 128, 256, 512, 512, 512} for the convolutional and deconvolutional layers in the six corresponding CBRR blocks of the encoder and the corresponding decoder modules. The experimental results indicate that there are only slight accuracy differences between the two models. Increasing the number of filters can only bring a slight improvement in accuracy to MSCFF, but it results in a significant increase in the use of computing resources. Our default setting of filter number can balance the model accuracy and computational efficiency.

## 3. Experimental data

The experimental datasets consisted of multiple types of optical satellite remote sensing imagery with different spatial resolutions. Four typical types of imagery were used for the model training and testing, i.e., Landsat-7 Enhanced Thematic Mapper Plus (ETM+) (30 m), Landsat-8 Operational Land Imager/Thermal Infrared Sensor (OLI/TIRS) (30 m), Gaofen-1 Wide-Field-View (WFV) (16 m), and high-resolution (HR) images (0.5–15 m) exported from Google Earth (Google Inc.). The details of the four datasets are given in Table 2. A total of 516 globally distributed scenes were selected for the experiments. Other types of optical imagery were used for the model performance validation by visual evaluation, i.e., Gaofen-1/2/4 Panchromatic and MultiSpectral Sensor (PMS), Ziyuan-3 multispectral (MUX), Sentinel-2A Multispectral Instrument (MSI), CBERS-04 P10, Landsat-5 Thematic Mapper (TM), and Huanjing-1A/1B charge-coupled device (CCD) imagery. To the best of our knowledge, this is the first time that such large and diverse datasets have been used for the validation of a newly developed cloud detection method. More details of the datasets are given in the following.

**Table 2** The four main experimental cloud and cloud shadow datasets. The *L7_Irish*, *L8_Biome*, and *GF1_WHU* datasets were acquired from public sources, while *HRC_WHU* was created in this study by experts in the field of remote sensing image interpretation. *Note that some of the images in the *L7_Irish* and *L8_Biome* datasets were removed from the experimental data because of intolerable errors in the labeling of cloud or cloud shadow. The total number of images denotes the images actually used.

| Name | Source | Resolution | Total images | Image size |
|---|---|---|---|---|
| L7_Irish (Scaramuzza et al., 2012) | Landsat-7 ETM+ | 30 m | 166* | ~7000×6000×7 |
| L8_Biome (Foga et al., 2017) | Landsat-8 OLI/TIRS | 30 m | 92* | ~7000×6000×9 |
| GF1_WHU (Li et al., 2017) | Gaofen-1 WFV | 16 m | 108 | ~17000×16000×4 |
| HRC_WHU (Ours) | Google Earth | 0.5–15 m | 150 | 1280×720×3 |

*L7_Irish*: The Landsat-7 cloud cover assessment validation dataset (Irish et al., 2006; Scaramuzza et al., 2012; USGS., 2016a) termed *L7_Irish* was derived by the U.S. Geological Survey (USGS) Earth Resources Observation and Science (EROS) Center. This collection of validation data contains 206 Landsat-7 ETM+ Level-1G scenes with associated cloud masks, in which only 45 masks are labeled for both cloud and cloud shadow. These scenes are stratified by latitude zone, with varying cloud conditions. The interpretation for the pixels in each mask includes cloud, thin cloud, and cloud shadow.

*L8_Biome*: The Landsat-8 cloud cover validation dataset (Foga et al., 2017; USGS., 2016b) termed *L8_Biome* was also created by USGS EROS. This collection of validation data contains 96 Landsat-8 OLI/TIRS terrain-corrected (Level-1T) scenes with associated manual cloud masks, in which only 32 masks are labeled for both cloud and cloud shadow. The numbers of scenes which have a percentage of cloud less than 35%, between 35% and 65%, and over 65% are all 32. These scenes are clustered into eight biomes, and each pixel in a scene is labeled as one of four classes, i.e., cloud, thin cloud, cloud shadow, and clear sky.

*GF1_WHU*: The Gaofen-1 cloud and cloud shadow cover validation dataset (Li et al., 2017) termed *GF1_WHU* in this paper was created by Wuhan University, and has been used for the performance evaluation of cloud detection methods on Gaofen-1 WFV imagery. The WFV imaging system onboard the Gaofen-1 satellite has a 16-m spatial resolution and four multispectral bands spanning the visible to the near-infrared spectral regions. This collection of validation data includes 108 Level-2A scenes collected from different global land-cover types with varying cloud conditions, and all the associated masks label both cloud and cloud shadow.

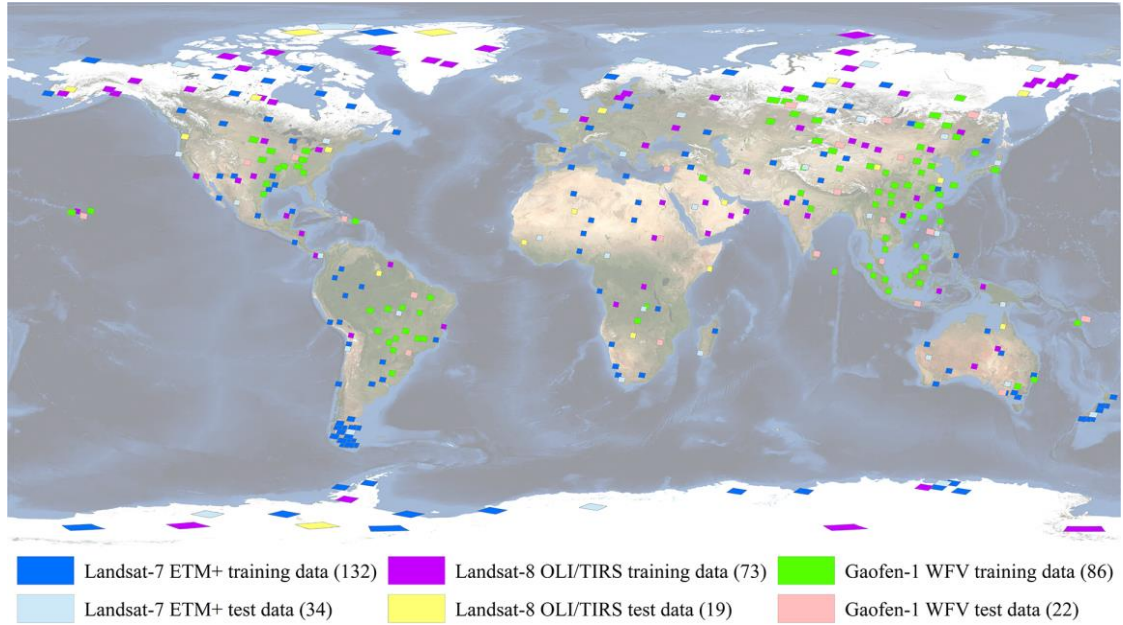

**Fig. 4.** Global distribution of the experimental data, including *L7_Irish*, *L8_Biome*, and *GF1_WHU*. The *HRC_WHU* data are not shown as they lack accurate geolocation information. The numbers in brackets denote the total number of images used for training or testing (base map credit: NASA Visible Earth).

*HRC_WHU*: We created a high-resolution cloud cover validation dataset in our study and termed it *HRC_WHU*. This was done because, as far as we know, there are no public high-resolution cloud detection datasets which can meet the training and testing requirements of deep models. The *HRC_WHU* data comprise 150 high-resolution images acquired with three RGB channels and a resolution varying from 0.5 to 15 m in different global regions. The images were collected from Google Earth (Google Inc.), in which satellite images, aerial photography, and geographic information system data are superimposed to map the Earth onto a 3D globe. The associated reference cloud masks were digitized by experts in the field of remote sensing image interpretation from Wuhan University. The established high-resolution cloud cover validation dataset has been made available online (http://sendimage.whu.edu.cn/en/mscff/). We believe that our public *HRC_WHU* dataset will contribute to the related cloud detection studies and communities, and will also benefit performance benchmarking of deep models in image classification tasks.

In the procedure of delineating the cloud mask for high-resolution imagery, we first stretched the cloudy image to the appropriate contrast in Adobe Photoshop. The lasso tool and magic wand tool were then alternately used to mark the locations of the clouds in the image. The manually labeled reference mask was finally created by assigning the pixel values of cloud and clear sky to 255 and 0, respectively. Note that a tolerance of 5–30 was set when using the magic wand tool, and the lasso tool was used to modify the areas that could not be correctly selected by the magic wand tool. As we did in a previous study (Li et al., 2017), the thin clouds were labeled as cloud if they were visually identifiable and the underlying surface could not be seen clearly. Considering that cloud shadows in high-resolution images are rare and hard to accurately select, only clouds were labeled in the reference masks.

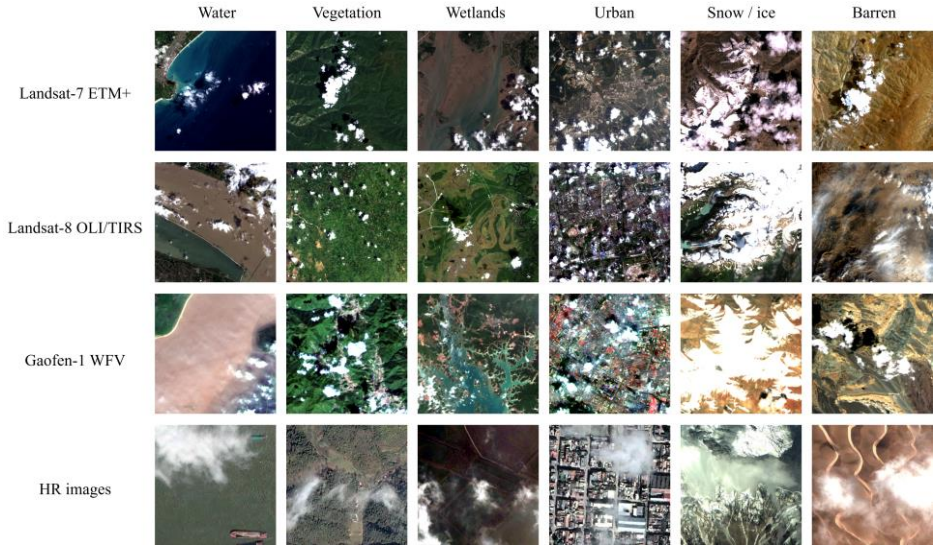

**Fig. 5.** Display of some of the experimental data in different land-cover types.

As shown in Fig. 4 and Fig. 5, the Landsat-7, Landsat-8, Gaofen-1, and high-resolution images are distributed in global regions with different land-cover types. It should be noted that some of the images and masks in the *L7_Irish* and *L8_Biome* datasets were removed from the data after careful visual inspection, because of intolerable cloud or cloud shadow labeling errors in the masks. Furthermore, we treated both cloud and thin cloud as cloud, and merged cloud shadow over land and water into the shadow class. Only cloud and cloud shadow were considered as targets for detection. We selected 80% of the scenes and masks in each dataset as training data, and the remainder were used as test data. The images used for model training and testing were completely independent. The training samples for the deep models were clipped from all the scenes of each training dataset, and all the channels in an image were used as the input. Specifically, 17991 Landsat-7 samples were clipped from 132 of 166 scenes, 10487 Landsat-8 samples were clipped from 73 of 92 scenes, and 59395 Gaofen-1 samples were clipped from 86 of 108 scenes. These samples were used as the training data, and all three types of training samples were clipped from corresponding scenes with a stride of 512 × 512. Meanwhile, the 6000 HR samples were generated from 120 of 150 scenes, in which the training samples were clipped partially overlapped to increase the number of training samples. The test scenes used in this study did not need to be pre-clipped, and were patch-by-patch processed by dividing the whole scenes into equal patches. We randomly selected test data from the whole of the dataset, to ensure that the numbers of images in different land-cover types were almost equal, as far as possible. The diversity and spatial independence of the training and test data allowed MSCFF and the compared methods to be more fully and accurately evaluated, as well as allowing a better validation of the generalization ability of the deep models.

More types of optical imagery which have no available cloud masks were also used for the method validation by visual evaluation, including Sentinel-2A MSI, Landsat-5 TM, Gaofen-1/2/4 PMS, Ziyuan-3 MUX, CBERS-04 P10, and Huanjing-1A/1B CCD imagery. All these types of imagery have visible and near-infrared bands. The pre-trained model was tested on the above types of imagery, without being trained on them, and the results are given in the application extension and limitations section.

## 4. Experimental results

### 4.1. Accuracy assessments and comparisons

In this subsection, we provide accuracy assessments and comparisons in terms of cloud and cloud shadow detection with the experimental data listed in Table 3. It should be noted that different models were trained for each dataset, since each sensor presents different spectral bands and spatial resolutions.

#### 4.1.1. Compared methods

The proposed MSCFF method was compared with representative rule-based and deep learning based cloud detection methods, both visually and quantitatively. Fmask, the multi-feature combined (MFC) method, and the progressive refinement scheme (PRS) are representative traditional rule-based methods for cloud detection in specific types of imagery. The Fmask algorithm (Zhu and Woodcock, 2012; Zhu et al., 2015) is a robust rule-based cloud and cloud shadow detection method developed for Landsat imagery. The MFC method (Li et al., 2017) combines multiple spectral and spatial features to implement threshold-based cloud and cloud shadow detection in Gaofen-1 WFV imagery. The PRS method (Zhang and Xiao, 2014) was designed for cloud detection in color aerial photographs.

DeepLab (Chen et al., 2018), which is a state-of-the-art deep segmentation network for natural images, and the deep convolutional network method developed by Zhan et al. (2017) (termed DCN in this paper), which was designed for cloud and snow detection tasks in satellite imagery, were selected as comparable deep learning architectures. We trained the DeepLab and DCN models from scratch with the same training data as we trained MSCFF.

We compared MSCFF with the above three rule-based methods and the two deep models on Landsat-7/8, Gaofen-1 WFV, and high-resolution imagery. Specifically, Fmask version 3.3 was used to generate masks for the Landsat-7 and Landsat-8 images. Cloud masks for the Gaofen-1 WFV and high-resolution imagery were produced by the public cloud detection tools of the MFC and PRS methods, respectively. The DeepLab model version used was the public *DeepLabv2_VGG16*, while the DCN model was constructed according to the provided details in the published paper. The first layers of the DeepLab and DCN models were both modified to accommodate the input of different channels. Note that only the cloud shadow detection results of Fmask, which are usually fragmentary, are dilated by one pixel to improve the performance in the accuracy assessment, while the cloud and cloud shadow results of MSCFF and the other compared methods are unchanged and without any mask dilation operation, in both the accuracy evaluation and the display of the experimental results.

**4.1.2. Accuracy metrics**

The accuracy assessment measures the agreements and differences between cloud and cloud shadow in the generated mask and the reference mask at the pixel level. For the cloud accuracy evaluation, cloud and non-cloud pixels were regarded as two categories, as were the cloud shadow and non-cloud shadow pixels for the accuracy evaluation of cloud shadow. The overall accuracy, recall, precision, mean intersection over union (mIoU), and F-score were used as the accuracy metrics to evaluate the quantitative performance of the compared methods. The five metrics can be calculated as follows:

$$Overall\ accuracy\ = \frac{TP+TN}{TP+TN+FP+FN} \qquad (2)$$

$$Recall\ = \frac{TP}{TP+FN} \qquad (3)$$

$$Precision = \frac{TP}{TP+FP} \quad (4)$$

$$mIoU = \frac{Intersection\ areas\ of\ detected\ and\ reference\ clouds}{Union\ areas\ of\ detected\ and\ reference\ clouds} \quad (5)$$

$$F1\ score = \frac{2}{\frac{1}{Recall} + \frac{1}{Precision}} \quad (6)$$

where $TP$ and $TN$ denote a correct prediction; and $FP$ and $FN$ denote an incorrect outcome in which the detected pixel is clear sky or cloud/cloud shadow, respectively. Note that the overall accuracy, mIoU, and F-score are comprehensive indicators, and a larger value means a higher accuracy, while recall and precision are related to the omission and commission errors, respectively.

### 4.1.3. Accuracy evaluation and analysis

We conducted accuracy tests on the four datasets for MSCFF and the compared methods. The detailed quantitative accuracy evaluation results for the cloud detection are listed in Table 4, from which we can see that MSCFF achieves the highest mIoU and F-score values among all the compared methods. Significant accuracy improvements can be observed for MSCFF over the traditional Fmask, MFC, and PRS methods, in terms of cloud detection. In addition, MSCFF also has accuracy advantages over the compared DeepLab and DCN methods.

**Table 4** Cloud detection accuracy comparisons on different types of test data. The bold values denote the highest accuracies for each dataset among the compared methods.

| Dataset | Method | Overall acc. | Recall | Precision | mIoU | F-score |
|---|---|---|---|---|---|---|
| Landsat-7 ETM+ (34 test scenes) | Fmask | 91.71% | 89.75% | 89.72% | 0.814 | 0.897 |
| | DeepLab | 90.19% | 87.31% | 88.25% | 0.782 | 0.878 |
| | DCN | 92.19% | 86.23% | **93.91%** | 0.817 | 0.899 |
| | MSCFF | **94.45%** | **92.51%** | 93.60% | **0.870** | **0.931** |
| Landsat-8 OLI/TIRS (19 test scenes) | Fmask | 89.59% | 93.01% | 85.80% | 0.806 | 0.893 |
| | DeepLab | 87.72% | 81.37% | 91.26% | 0.755 | 0.860 |
| | DCN | 92.37% | 87.27% | **95.96%** | 0.842 | 0.914 |
| | MSCFF | **94.96%** | **93.93%** | 95.05% | **0.895** | **0.945** |
| Gaofen-1 WFV (22 test scenes) | MFC | 95.96% | 91.46% | 94.35% | 0.867 | 0.929 |
| | DeepLab | 91.29% | 73.80% | 94.84% | 0.710 | 0.830 |
| | DCN | 96.28% | 90.50% | 96.38% | 0.875 | 0.933 |
| | MSCFF | **96.83%** | **92.05%** | **96.81%** | **0.893** | **0.944** |
| HR images (30 test scenes) | PRS | 86.66% | 77.00% | 87.83% | 0.696 | 0.821 |
| | DeepLab | 93.83% | 92.66% | 91.84% | 0.856 | 0.922 |
| | DCN | 94.91% | 92.02% | 94.97% | 0.877 | 0.935 |
| | MSCFF | **95.99%** | **94.41%** | **95.43%** | **0.903** | **0.949** |

According to our visual inspection, for each test result with Landsat imagery, MSCFF and the compared Fmask method both produce accurate masks. The Fmask method is effective in capturing thin clouds, and shows a robust performance under most conditions. However, Fmask easily mistakes bright surfaces as clouds, especially in snow/ice covered areas, coastal areas, bright water body areas, etc. MSCFF achieves overall accuracies for the cloud detection of 94.45% and 94.96% for the 34 tested Landsat-7 scenes and 19 tested Landsat-8 scenes, respectively, while the overall accuracies for the cloud detection of Fmask are 91.71% and 89.59%. The mIoU cloud accuracies of DCN are higher than those of Fmask and DeepLab, but are lower than those of MSCFF for the Landsat-7 and Landsat-8 imagery, because of omitting more thin clouds around cloud boundaries. Given the very few ground-truth images that contain cloud shadow labels in the *L7_Irish* and *L8_Biome* datasets, and the fact that the percentages of cloud shadow in the tested Landsat images are relatively low, the cloud shadow detection results of Fmask are regarded as correct cloud shadow labels and are used in the Landsat training images that contain only cloud labels. It should be noted that the cloud shadow accuracies for the Landsat images can only be roughly evaluated, and were computed over ten tested Landsat-7 scenes and five tested Landsat-8 scenes that both originally labeled cloud and cloud shadow in the test data shown in Fig. 4. MSCFF achieves the highest mIoU and F-score values among the compared methods in both the tested Landsat-7 and Landsat-8 scenes. Specifically, taking the five tested Landsat-8 scenes as an example, the cloud shadow recall and precision accuracies of MSCFF are 63.32% and 70.32%, respectively, while Fmask achieves cloud shadow precision accuracies of 55.41% and 44.65%, which indicates that MSCFF performs better than Fmask in cloud shadow detection, although partial cloud shadow training labels are derived from the results of Fmask. However, more tested Landsat scenes that contain correct cloud shadow labels are needed to evaluate the cloud shadow detection accuracy of the different methods more accurately. Fig. 6 gives cloud and cloud shadow detection examples of the above-mentioned cases on Landsat-7 ETM+ and Landsat-8 OLI/TIRS imagery, from which we can observe that MSCFF is more effective at distinguishing bright surfaces from clouds than Fmask and the other compared methods, and the cloud detection result of MSCFF is closer to the ground truth. The cloud shadow detection result of MSCFF is visually satisfactory, and is more accurate than that of the compared methods.

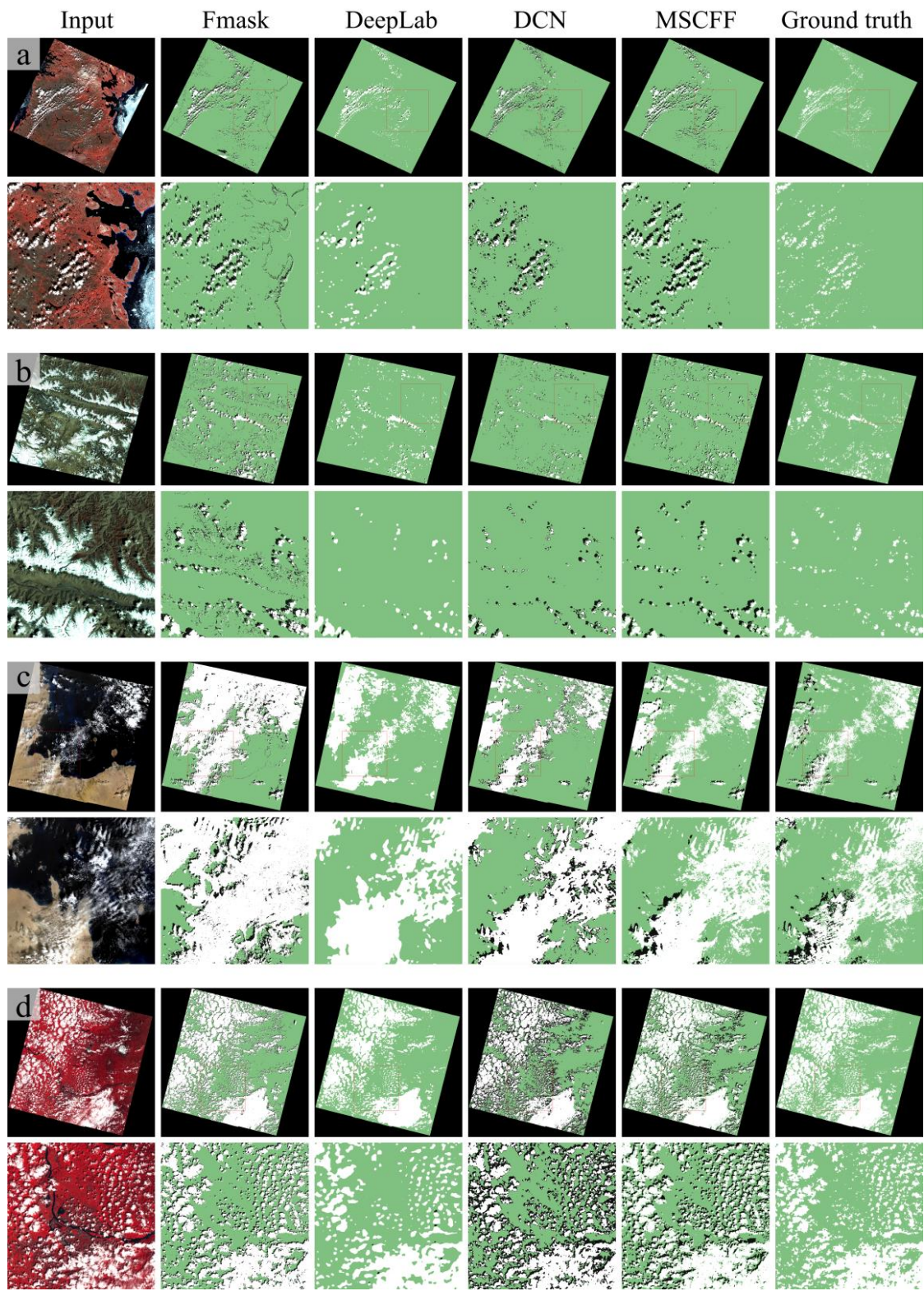

**Fig. 6.** Examples of cloud and cloud shadow detection results for Landsat imagery by the different methods. (a) Vegetation area (scene ID: L7_p030r012_20010811). (b) Snow/ice area (scene ID: L7_p147r035_20010511). (c) Water area (scene ID: LC8_p162r043_2014072). (d) Vegetation area (scene ID: LC8_p046r028_2014171). Noted that ground truths for images (a), (b), and (d) are provided without labeling cloud shadow. The shown input images are composited with NIR-R-G channels, and the white, black, and green colors in the masks of valid areas denote cloud, cloud shadow, and clear sky, respectively.

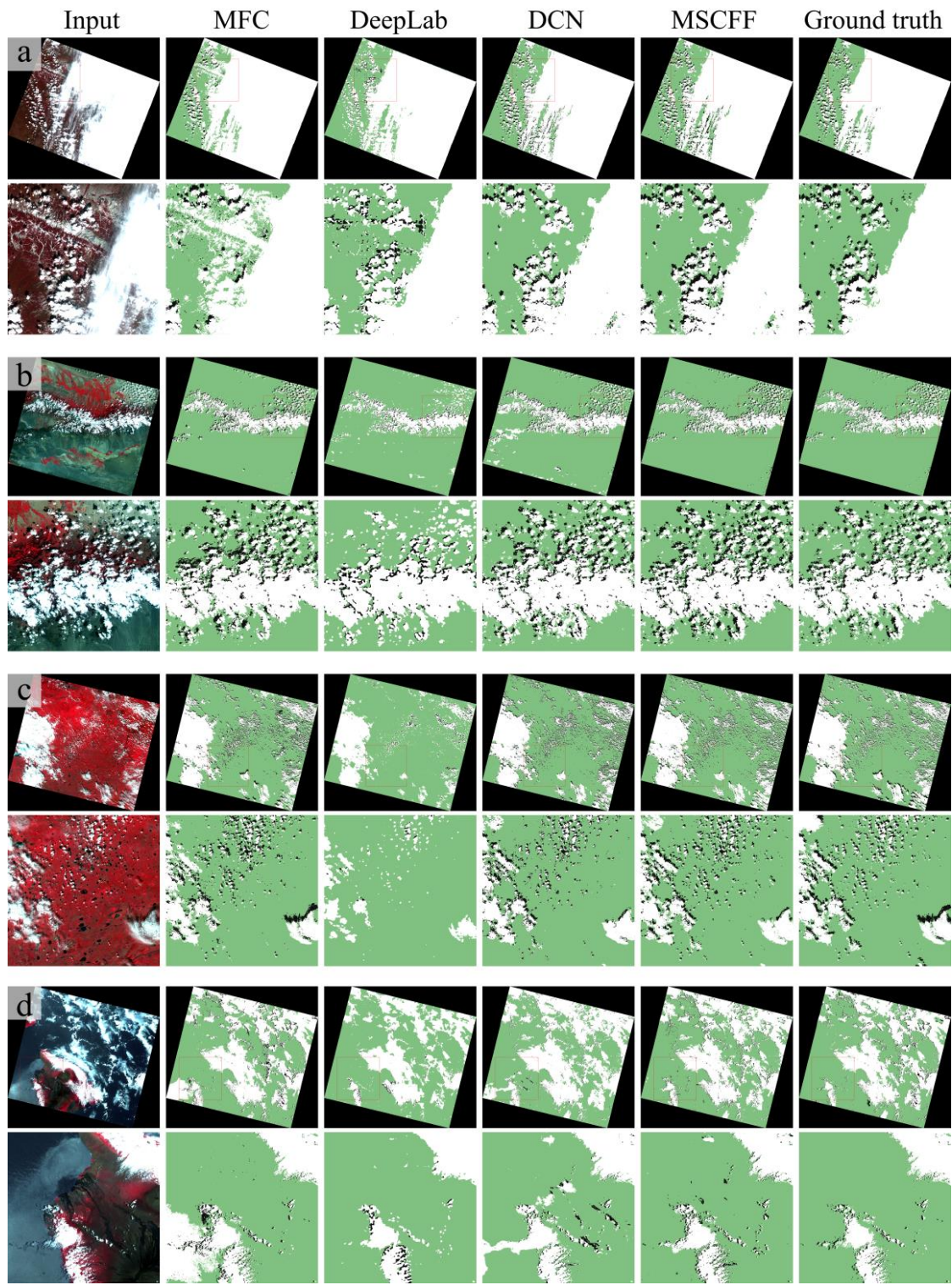

**Fig. 7.** Examples of cloud and cloud shadow detection results for Gaofen-1 WFV imagery by the different methods. (a) Snow/ice area (scene ID: E132.4_N53.2_20160507). (b) Barren area (scene ID: E89.3_N43.4_20130714). (c) Vegetation area (scene ID: E73.7_N56.3_20130712). (d) Water area (scene ID: W155.2_N20.2_20160721). The shown input images are composited with NIR-R-G channels, and the white, black, and green colors in the masks of valid areas denote cloud, cloud shadow, and clear sky, respectively.

Cloud and cloud shadow detection examples on Gaofen-1 WFV imagery are shown in Fig. 7, in which it can be seen that MSCFF generates more accurate cloud masks than MFC in areas of bright surfaces, including bright water bodies and snow, and thus MSCFF achieves an F-score cloud accuracy of 0.944, which is higher than the 0.929 of MFC. The F-score cloud accuracies of DeepLab and DCN are 0.830 and 0.933, respectively. The cloud masks generated by DeepLab are rough and have obvious mistakes in local regions, which results in DeepLab having a lower cloud recall accuracy, while the cloud detection results of DCN are generally satisfactory, but are less accurate than those of MSCFF, especially in large areas of bright surfaces. As for the evaluation of cloud shadow detection, due to the fact that there are fewer cloud shadows than clouds in the training samples, the cloud shadow accuracy of the deep models is not as high as for cloud, according to our evaluation. The average cloud shadow recall and precision accuracies of MSCFF are 81.35% and 76.16%, respectively, among the 22 tested Gaofen-1 WFV scenes, which are much higher than those of DCN with 71.81% and 60.21%, respectively, because of the more accurate detection of cloud shadow objects and the reduced commission of dark objects such as water bodies and terrain shadows. It should be noted that the low percentage of 4.45% cloud shadow pixels in the Gaofen-1 test scenes may result in bias in the evaluation results of the cloud shadow detection.

The tested high-resolution images were acquired with different spatial resolutions and land covers. The deep models in this paper were trained to only detect clouds in high-resolution images, due to the fact that there were very few cloud shadows in the training samples. It should be noted that the PRS method can only be relatively compared with the deep models, as it is designed for cloud detection in color aerial images, and three-channel aerial images are not exactly the same as the high-resolution images acquired from Google Earth. Fig. 8 shows the cloud detection results of the different methods. We can observe that PRS only detects the thick clouds, and misclassifies parts of the snow as clouds, and thus only 77.00% recall accuracy and 87.83% precision accuracy are achieved. The cloud detection results of DeepLab are rough and lose details in the cloud boundaries, while DCN produces more refined cloud masks, and the clouds captured by MSCFF are the most accurate among the compared methods. On the tested high-resolution images, MSCFF acquires an mIoU score of 0.903, while DeepLab and DCN achieve 0.856 and 0.877, respectively, which indicates that MSCFF is a more effective model for cloud detection in high-resolution images.

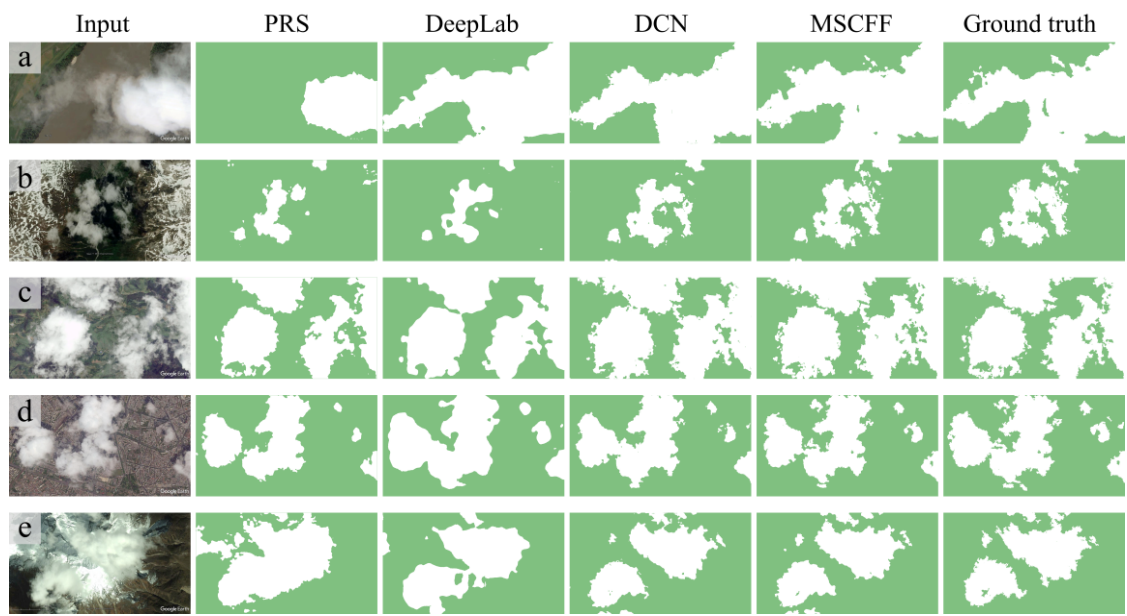

**Fig. 8.** Examples of cloud and cloud detection results for Google Earth imagery by the different methods. (a) Water area. (b) Bright surface area. (c) Vegetation area. (d) Urban area. (e) Snow/ice area. The shown images are composited with RGB channels, and the white color in the masks denotes cloud.

In addition, to more comprehensively evaluate the results of the different methods, the cloud detection accuracy under different land-cover types was also assessed. Specifically, a global MODIS Land Cover Type Climate Modeling Grid (CMG) product (Short Name: MCD12C1) (Fallis, 2013) of 0.05-degree resolution was applied to classify the validation datasets into eight classes of water, forest, shrubland, grass/crops, wetlands, urban, snow/ice, and barren, according to the International Geosphere Biosphere Programme (IGBP) (Loveland et al., 2000) classification scheme. Fig. 9 shows the cloud detection accuracies of the different methods under the different land covers in Landsat and Gaofen-1 imagery. We can observe that MSCFF achieves an obvious improvement compared to Fmask in snow/ice covered areas on the 53 tested Landsat-7 and Landsat-8 images, from 87.05% to 90.20%, and shows an improvement in urban areas from 86.16% to 91.72%. On the 22 tested Gaofen-1 WFV images, MSCFF and MFC both acquire high accuracies in the vegetation, urban, and wetland areas, and the pixel-level average overall accuracy of the cloud detection in the snow-covered areas is significantly improved from 65.39% with MFC to 92.17% with MSCFF. Among the compared methods and test images, MSCFF achieves the highest accuracies over most of the land-cover types.

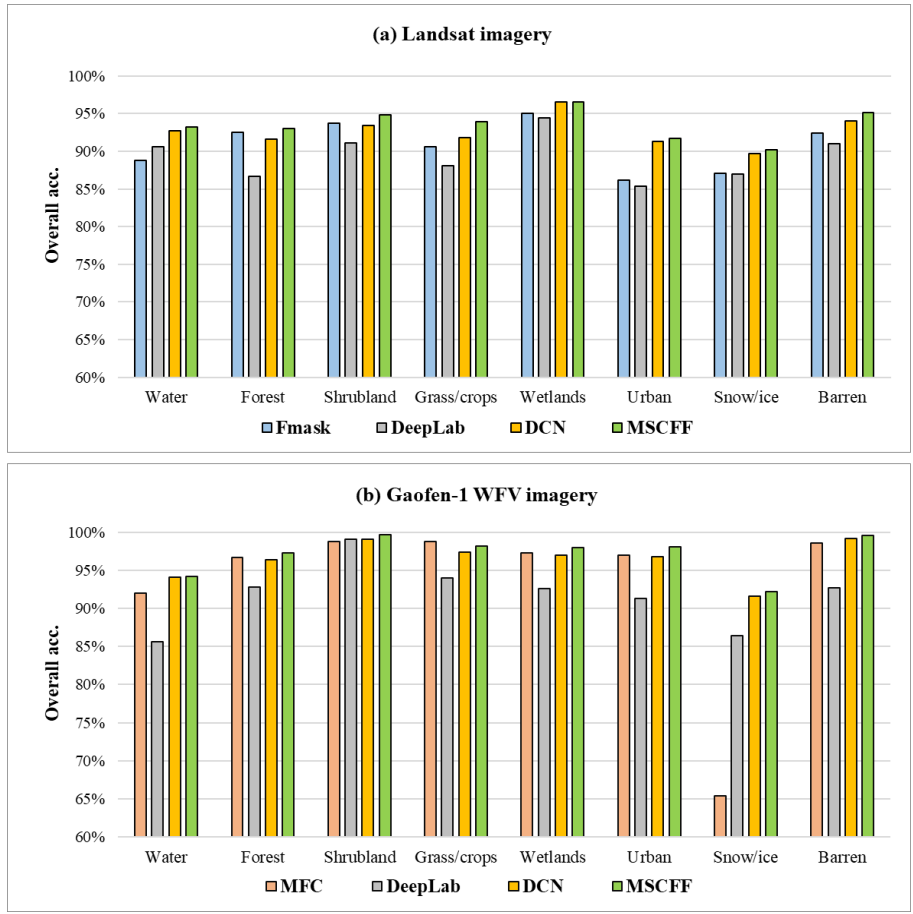

**Fig. 9.** Cloud detection accuracy of the different methods in different land-cover types. (a) Average overall accuracy of the cloud detection for Landsat-7 and Landsat-8 imagery. (b) Average overall accuracy of the cloud detection for Gaofen-1 WFV imagery.

In general, the traditional cloud detection methods are able to capture most of the clouds in satellite imagery. Benefiting from the abundant spectral information in Landsat imagery when compared to Gaofen-1 WFV and high-resolution imagery, the Fmask method is sensitive to the thin clouds in Landsat imagery. However, it is clear that Fmask, MFC, and PRS easily mistake the bright surfaces as clouds. Furthermore, PRS has the shortcoming of omitting the thin clouds around cloud boundaries in high-resolution images. The deep models of DeepLab, DCN, and MSCFF are more effective than the traditional methods in coping with cases of bright surfaces. Among the three deep learning based methods, the results of DeepLab are found to be not fine enough in the cloud boundaries, DCN misclassifies some bright surfaces and misses some small clouds in local areas, while MSCFF obtains fewer errors than DeepLab and DCN in the masks and acquires the highest accuracies. There are two main reasons for MSCFF producing more accurate masks than the compared methods. On the one hand, the use of the symmetric encoder-decoder module in MSCFF can help to fuse the feature hierarchy to combine the shallow appearance information and the deep semantic information. The multi-scale convolutional feature fusion gives MSCFF a stronger ability to distinguish clouds from bright non-cloud objects. Residual learning and dilated convolutions are additionally applied in MSCFF to achieve better optimal convergence in the model training and increase the receptive field size, respectively, without losing detail information, thus boosting the model accuracy. On the other hand, the proposed network is trained on globally distributed datasets which consist of multiple land-cover types, so the diversity of the training data makes MSCFF effective, and gives it a strong capability to cope with complex surfaces.

In addition to the model architecture, the input spectral bands are a key factor influencing the performance of the cloud detection methods. In this study, in order to compare the performances of MSCFF under the circumstances of different inputs, the inputs of RGB bands and NIR-R-G-B bands for MSCFF were both evaluated. The results are given in Table 5, in which $MSCFF_{RGB}$ and $MSCFF_{NRGB}$ denote the network trained on the Landsat-8 images with only visible bands and near-infrared/visible bands, respectively. The accuracy evaluation results for MSCFF in the different input situations indicate that the input of all the available bands in Landsat-8 images achieves the highest overall accuracies in the tested images, and the addition of the near-infrared band improves the cloud detection accuracy of MSCFF. In addition, the different band settings in the different types of imagery necessitate cloud detection algorithms both with and without infrared and thermal data to generate masks, and it has been proven that MSCFF is capable of dealing with different band settings.

**Table 5** Accuracy tests for MSCFF on Landsat-8 imagery with different inputs.

| Test dataset | Method | Overall acc. | Recall | Precision | mIoU | F-score |
|---|---|---|---|---|---|---|
| | $MSCFF_{RGB}$ | 93.50% | 91.76% | 90.63% | 0.866 | 0.912 |
| Landsat-8 OLI/TIRS | $MSCFF_{NRGB}$ | 93.94% | 94.52% | 88.08% | 0.878 | 0.926 |
| | MSCFF | 94.96% | 93.93% | 95.05% | 0.895 | 0.945 |

The accuracy of cloud detection is also related to the spatial resolution of the imagery. As the spatial resolutions of different types of imagery change, there is no obvious change in the spectral characteristics of cloud and cloud shadow. However, the spatial features of some bright surfaces, such as buildings, will change as the spatial resolution changes. Therefore, scale problems may exist when applying the pre-trained model to another type of imagery. In order to better deal with the scale problems and make the model adaptive to changes of the spatial resolution, images of different resolutions and scales should be included in the training data, to give the network a stronger ability to cope with imagery of different resolutions.

**4.2. Efficiency**

Taking a Landsat-8 image as an example, we tested the efficiency of the different methods. It should be noted that Fmask was tested under CPU mode with an Intel Core i7-7700K, while DeepLab, DCN, and MSCFF were tested with NVIDIA Titan Xp GPU support. In our implementation, MSCFF takes less than a minute in GPU mode to process a whole Landsat-8 image under the MatConvNet environment. Table 6 lists the test results, from which we can see that the deep learning based methods of DeepLab, DCN, and MSCFF cost less time than the traditional rule-based Fmask method under the GPU acceleration. Note that the time cost of Fmask is related to the image size and cloud distribution, which influence the speed of the cloud and cloud shadow matching, while the time costs of the deep learning based methods depend on the image size only. Although the time efficiency of MSCFF is a bit lower than that of DeepLab, the accuracy of MSCFF is much more satisfactory. The model size of MSCFF is much smaller than that of DeepLab and DCN, which means that MSCFF achieves the highest accuracy, but with significantly fewer parameters, thus indicating the effectiveness of MSCFF.

**Table 6** Efficiency comparison of the different methods. *The Fmask method was tested on an Intel Core i7-7700K CPU only, while the other deep learning based methods were tested with support from an NVIDIA Titan Xp GPU.

| Test image | Method | Time cost (seconds) | Model size (Mb) |
|---|---|---|---|
| Landsat-8 OLI/TIRS | Fmask | 113.8* | N/A |
| Scene ID: LC8_p16r50_2014041 | DeepLab | 42.9 | 267 |
| | DCN | 112.1 | 461 |
| Size: 7721×7541×8 | MSCFF | 47.4 | 10 |

**5. Application extension and limitations**

Extending the application of the model to other images is investigated in this section. Furthermore, the limitations of the proposed MSCFF method are discussed.

**5.1. Application extension**

The pre-trained MSCFF model can also be used to extract clouds for other types of satellite images with similar spectral settings, without needing any parameter adjustment. Considering that imagery which has a large digital number range may have low image contrast after maximum normalization, before the cloudy image is fed into the MSCFF model, an image stretching preprocessing step should be considered to make sure that the input image has natural visual effects. MSCFF is not as sensitive as the traditional rule-based methods to the light condition changes in the image, but the input of a contrast- and color-balanced image for MSCFF will help it to generate a more accurate mask. In our implementation, for an image that has low contrast, linear transformation is conducted before we input the image to MSCFF. For images which have inner spectral differences, the linear transform coefficients for each band in an image should be different, to better simulate actual cases.

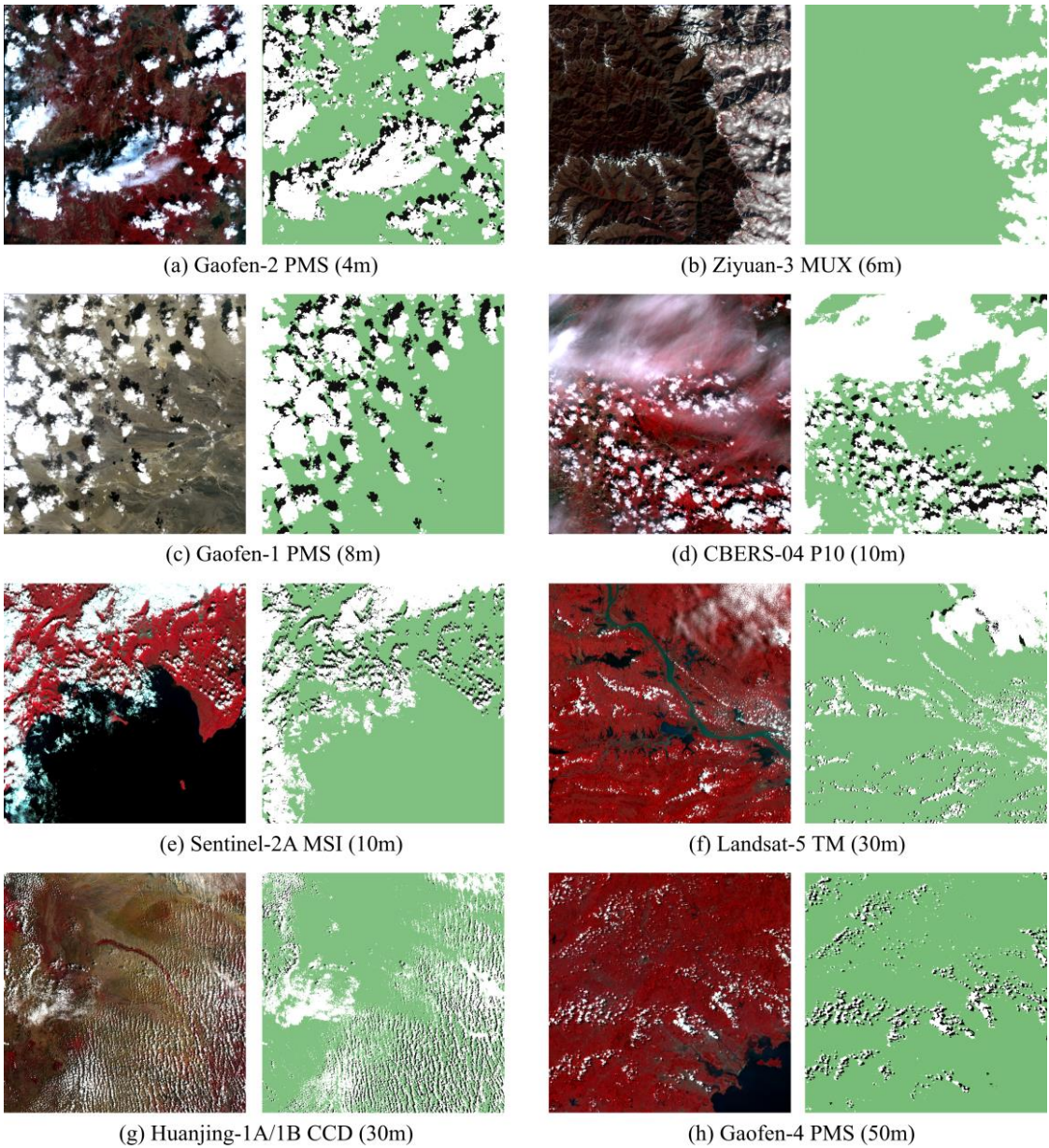

(a) Gaofen-2 PMS (4m)  (b) Ziyuan-3 MUX (6m)
(c) Gaofen-1 PMS (8m)  (d) CBERS-04 P10 (10m)
(e) Sentinel-2A MSI (10m)  (f) Landsat-5 TM (30m)
(g) Huanjing-1A/1B CCD (30m)  (h) Gaofen-4 PMS (50m)

**Fig. 10.** Cloud detection examples of MSCFF in multiple types of images of different land covers. (a) Scene ID: GF2_PMS2_E99.2_N24.5_20160320. (b) Scene ID: ZY3_MUX_E103.7_N32.3_20140109. (c) Scene ID: GF1_PMS1_E90.4_N41.6_20150419. (d) Scene ID: CB04_P10_373_52_20160822. (e) Scene ID: S2A_MSIL1C_20180813_N0206_R102. (f) Scene ID: LT5_122_039_2004181. (g) Scene ID: HJ1A_CCD1_88_120_20141107. (h) Scene ID: GF4_PMS_E114.7_N22.9_20160723.

The pre-trained MSCFF model, which was trained on the globally distributed Gaofen-1 WFV dataset, was applied to more satellite images with resolution ranges from 4 m to 50 m, i.e., Gaofen-2 PMS (4 m), Ziyuan-3 MUX (6 m), Gaofen-1 PMS (8 m), Sentinel-2A MSI (10 m), Landsat-5 TM (30 m), and Gaofen-4 PMS (50 m), as shown in Fig. 10. It can be seen that MSCFF generates visually satisfactory cloud masks for the multiple types of imagery. To the best of our knowledge, this is the first time that a single cloud detection method has been developed that can deal with so many types of satellite images with different resolutions.

Model fine-tuning also works for transferring the pre-trained model, to make it perform better for new types of images, by re-training the model with a small learning rate and the training data of a new image type, which may have few labels.

**5.2. Limitations**

Although the proposed MSCFF method can achieve a high cloud detection accuracy in different complex surface conditions, there are still a few cases that MSCFF may fail to cope with. Limited by the receptive field size of the MSCFF architecture and the size of the input image patch, in the absence of a thermal infrared band as input, MSCFF may misclassify central regions of large-area bright objects such as snow, because the core regions of bright surfaces are similar to clouds, and the deep model cannot extract discriminative features in the patch-wise processing, as the patch size is smaller than the size of the object. This case is rare in actual images, but may occur in high-altitude snow/ice covered imagery. Fig. 11(a) gives an example of such a case in Gaofen-1 WFV imagery, which only contains visible and near-infrared channels. Expanding the receptive field of MSCFF by increasing the size of the input image patch or the depth of the network can help to solve such a problem. However, this requires computing device support and increases the complexity of the network.

The performances of deep models are also closely related to the training samples. Without enough cloud shadow training samples, MSCFF may omit some so-called bright cloud shadow, especially in very high resolution imagery, as shown in Fig. 11(b), as cloud shadows are less evident in high-resolution images than in medium- and low-resolution images. In addition, cloud shadows in high-resolution images are easily confused with building shadows. The classical methods try to associate clouds and cloud shadows by using the scene solar and viewing angles, which makes cloud shadow detection for high-resolution imagery a challenging task. Another issue is the scale problem between images of different resolutions, and the accompanying relationships between cloud and cloud shadow, which should be learned from images of multiple resolutions, as they vary with the change of the spatial resolution. Considering that the model which is used to detect cloud shadows in this paper is trained on medium-resolution images, it is not recommended to extendedly apply the pre-trained model to high-resolution images for cloud shadow detection. To overcome this shortcoming, more high-resolution training samples which contain sufficient cloud shadow labels are needed.

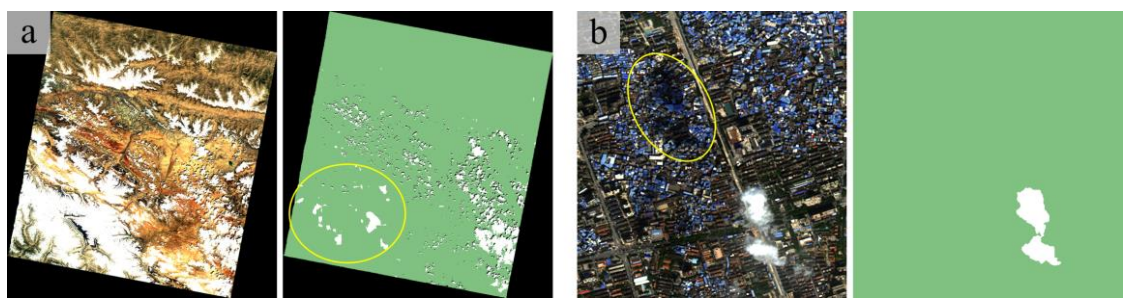

**Fig. 11**. Failed cases of MSCFF cloud and cloud shadow detection. (a) Mistaken identification of large-area snow as clouds in Gaofen-1 WFV imagery. (b) Missed cloud shadows in high-resolution TripleSat imagery.

## 6. Conclusions

In general, the traditional rule-based methods of cloud and cloud shadow detection perform well in medium- and low-resolution imagery, which has relatively abundant spectral information. However, with the increase of the image spatial resolution and the decrease of the available spectral channels, the cloud and cloud shadow detection accuracy generally decreases, because of the easily confused bright land surfaces and thin clouds that are difficult to accurately capture. In this case, the combined use of multiple features (spectral, spatial, and temporal features) can help to slow the accuracy degradation process. Nevertheless, it is difficult to accurately describe the characteristics of a target by a limited number of features under a complex surface. In this paper, the combination of multi-scale convolutional features makes the accurate definition of the target as close to perfect as possible, thus effectively reducing the negative influences on cloud and cloud shadow detection due to the increased spatial resolution and reduced spectral bands.

In this paper, we have proposed the multi-scale convolutional feature fusion (MSCFF) method to implement cloud detection for medium- and high-resolution satellite images of different sensors. The experimental results indicate that the feature fusion module and the residual network units both help to improve the model performance. MSCFF achieved a higher accuracy than the compared methods on different datasets, especially in bright surface covered areas. In addition, MSCFF was applied to extract clouds and cloud shadows for more than 10 types of satellite imagery. The effectiveness of MSCFF means that it shows great promise for practical application with multiple types of satellite imagery.

In our future study, we will attempt to improve the cloud shadow detection accuracy, as it is more difficult than cloud detection, even for the advanced methods, in which cloud shadows are easily confused with dark objects such as mountain shadows, building shadows, and water bodies, especially for images with limited spectral information. In addition, limited by the size of the input image patch and the receptive field size of CNNs, large-area cloud shadows may be identified fragmentarily. In this case, a combination of a CNN and object-based image analysis (OBIA) may help to improve the accuracy. Furthermore, to better overcome the weakness of the current models, which have a low accuracy for classes of insufficient training samples, such as cloud shadow, a better training strategy considering the class frequencies of the training samples will be investigated. In addition, we will generalize the MSCFF method to more types of imagery, and explore the possibility of using a single model to extract cloud and cloud shadow from multiple types of imagery.

**Acknowledgments**

This research was supported by the National Key Research and Development Program of China (No. 2017YFA0604402); the National Natural Science Foundation of China (NSFC) under Grant Nos. 61671334 and 41601357; the China Land Surveying and Planning Institute (CLSPI) under the development of cloud detection software for multi-source satellite images in national land resources investigation. We would like to thank the NVIDIA Corporation for the GPU donation, the USGS for their public cloud cover validation datasets, the authors of the DCN method for providing the experimental results, and Dr. Qing Zhang for providing the tool of the PRS method. This work has partly benefited from the public code of the Fmask method. The authors of this paper would also like to thanks the editors and the three anonymous reviewers for providing the valuable comments, which helped to greatly improve the manuscript.

**References**

Badrinarayanan, V., Kendall, A., Cipolla, R., 2017. SegNet: A Deep Convolutional Encoder-Decoder Architecture for Image Segmentation. IEEE Trans. Pattern Anal. Mach. Intell. 39, 2481–2495. https://doi.org/10.1109/TPAMI.2016.2644615

Bai, T., Li, D., Sun, K., Chen, Y., Li, W., 2016. Cloud detection for high-resolution satellite imagery using machine learning and multi-feature fusion. Remote Sens. 8, 1–21. https://doi.org/10.3390/rs8090715

Bian, J., Li, A., Liu, Q., Huang, C., 2016. Cloud and snow discrimination for CCD images of HJ-1A/B constellation based on spectral signature and spatio-temporal context. Remote Sens. 8. https://doi.org/10.3390/rs8010031

Bottou, L., 2010. Large-Scale Machine Learning with Stochastic Gradient Descent. Proc. COMPSTAT'2010. https://doi.org/10.1007/978-3-7908-2604-3_16

Braaten, J.D., Cohen, W.B., Yang, Z., 2015. Automated cloud and cloud shadow identification in Landsat MSS imagery for temperate ecosystems. Remote Sens. Environ. 169, 128–138. https://doi.org/10.1016/j.rse.2015.08.006

Chen, L.C., Papandreou, G., Kokkinos, I., Murphy, K., Yuille, A.L., 2018. DeepLab: Semantic Image Segmentation with Deep Convolutional Nets, Atrous Convolution, and Fully Connected CRFs. IEEE Trans. Pattern Anal. Mach. Intell. 40, 834–848. https://doi.org/10.1109/TPAMI.2017.2699184

Choi, H., Bindschadler, R., 2004. Cloud detection in Landsat imagery of ice sheets using shadow matching technique and automatic normalized difference snow index threshold value decision. Remote Sens. Environ. 91, 237–242. https://doi.org/10.1016/j.rse.2004.03.007

Di Vittorio, A. V., Emery, W.J., 2002. An automated, dynamic threshold cloud-masking algorithm for daytime AVHRR images over land. IEEE Trans. Geosci. Remote Sens. 40, 1682–1694. https://doi.org/10.1109/TGRS.2002.802455

Dumoulin, V., Visin, F., 2018. A guide to convolution arithmetic for deep learning. 1–31. arXiv preprint arXiv:1603.07285

Fallis, A.., 2013. User Guide for the MODIS Land Cover Type Product (MCD12Q1). https://lpdaac.usgs.gov/dataset_discovery/modis/modis_products_table/mcd12q1

Fisher, A., 2014. Cloud and cloud-shadow detection in SPOT5 HRG imagery with automated morphological feature extraction. Remote Sens. 6, 776–800. https://doi.org/10.3390/rs6010776

Foga, S., Scaramuzza, P.L., Guo, S., Zhu, Z., Dilley, R.D., Beckmann, T., Schmidt, G.L., Dwyer, J.L., Joseph Hughes, M., Laue, B., 2017. Cloud detection algorithm comparison and validation for operational Landsat33